\pdfoutput=1

\documentclass[11pt]{article}
\usepackage{booktabs}
\usepackage{longtable}
\usepackage{amsmath}
\usepackage{booktabs}

\usepackage[preprint]{acl}

\usepackage{times}
\usepackage{latexsym}
\usepackage[utf8]{inputenc}
\usepackage[T2A]{fontenc}
\usepackage[russian,english]{babel}
\usepackage{multirow}
\usepackage{adjustbox, booktabs, multirow, multicol, amssymb, url}

\usepackage[T1]{fontenc}

\usepackage[utf8]{inputenc}

\usepackage{microtype}

\usepackage{inconsolata}

\usepackage{graphicx}
\usepackage[dvipsnames]{xcolor}
\usepackage[many]{tcolorbox}
\usepackage{xcolor}
\usepackage{caption}
\usepackage{float}
\usepackage{pifont}        
\usepackage{xcolor}        
\usepackage{booktabs}      

\tcbuselibrary{listings}
\usepackage{color}

\title{\textit{Do Multi-Agents Solve Better Than Single?} Evaluating Agentic Frameworks for Diagram-Grounded Geometry Problem Solving and Reasoning}

\author{
 \textbf{Mahbub E Sobhani\textsuperscript{1, 2}}\thanks{Equal Contribution},
 \textbf{Md. Faiyaz Abdullah Sayeedi\textsuperscript{2, 3}}\textsuperscript{*},
 \textbf{Mohammad Nehad Alam\textsuperscript{1}},\\
 \textbf{Proma Hossain Progga\textsuperscript{4}},
 \textbf{Swakkhar Shatabda\textsuperscript{1}}
\\
\\
\textsuperscript{1} BRAC University, 
 \textsuperscript{2} United International University, \\
 \textsuperscript{3} Center for Computational \& Data Sciences, Independent University, Bangladesh, \\
\textsuperscript{4} Spectrum Software \& Consulting Ltd
\\
 \small{
   \textbf{Correspondence:} \href{swakkhar.shatabda@bracu.ac.bd}{swakkhar.shatabda@bracu.ac.bd}
 }
}

\newcommand{\methodname}{Interpreter-Solver}

\begin{document}
\maketitle
\begin{abstract}

Diagram-grounded geometry problem solving is a critical benchmark for multimodal large language models (MLLMs), yet the benefits of multi-agent design over single-agent remain unclear. We systematically compare single-agent and multi-agent pipelines on four visual math benchmarks: Geometry3K, MathVerse, OlympiadBench, and We-Math. For open-source models, multi-agent consistently improves performance. For example, Qwen-2.5-VL (7B) gains +6.8 points and Qwen-2.5-VL (32B) gains +3.3 on Geometry3K, and both Qwen-2.5-VL variants see further gains on OlympiadBench and We-Math. In contrast, the closed-source Gemini-2.0-Flash generally performs better in single-agent mode on classic benchmarks, while multi-agent yields only modest improvements on the newer We-Math dataset. These findings show that multi-agent pipelines provide clear benefits for open-source models and can assist strong proprietary systems on newer, less familiar benchmarks, but agentic decomposition is not universally optimal. All code, data, and reasoning files are available at \url{https://github.com/faiyazabdullah/Interpreter-Solver}
\end{abstract}

\section{Introduction}\label{intro}

Solving geometry problems from diagrams and natural language remains a challenging task requiring both visual understanding and symbolic reasoning. Prior studies highlight the effectiveness of neuro-symbolic frameworks, where transformer-based models such as BART \citep{lewis-etal-2020-bart}, LLaVA \citep{liu2024improved}, and Qwen \citep{bai2025qwen25vltechnicalreport} learn joint embeddings to bridge perception and reasoning. Theorem-based solvers \citep{lu-etal-2021-inter} and architectural refinements like enhanced multimodal alignment \citep{gao2023g} further improved performance. However, fine-tuning remains resource-intensive \citep{gao2023g}, theorem-prediction approaches suffer from overestimation bias \citep{peng-etal-2023-geodrl}, and methods such as AutoGPS \citep{ping2025autogps} depend on manual formalization.  


These limitations raise a key question: \textit{Can vision–language models (VLMs) and large language models (LLMs) solve geometry problems collaboratively in a zero-shot setting without task-specific supervision?} To address this, we systematically compare \textbf{single-agent} and \textbf{multi-agent} pipelines. Our findings show that multi-agent decomposition generally benefits open-source models, while advanced closed-source systems often perform better in single-agent mode, suggesting that decomposition is not universally optimal. The contributions of this paper are as follows: 
\begin{itemize}
    \item We provide the systematic comparison of single-agent and multi-agent pipelines for geometry problem solving.  
    \item We benchmark both paradigms on four benchmarks, Geometry3K, MathVerse, OlympiadBench, and We-Math, finding consistent multi-agent gains for open-source models and mostly single-agent advantages for strong closed-source models, with modest multi-agent gains on the newer benchmarks where contamination is less likely.
    \item We situate our approach against prior baselines, achieving new state-of-the-art results in zero-shot settings with significantly fewer parameters.  
\end{itemize}

\section{Related Work}\label{back}

Solving geometrical problems requires combining symbolic reasoning with multimodal comprehension of diagrams and text, making it a challenging benchmark for both symbolic and neural approaches. Symbolic methods, such as the parser-based solver of \citet{lu-etal-2021-inter}, achieve interpretable results but depend on handcrafted rules. With the advent of multimodal large language models (MMLMs), transformer-based systems like Qwen \citep{bai2025qwen25vltechnicalreport}, PaLI \citep{chen2023palijointlyscaledmultilinguallanguageimage}, LLaVA \citep{NEURIPS2023_6dcf277e}, Gemini \citep{comanici2025gemini}, and GPT \citep{achiam2023gpt} have been widely adopted. Extensions include Progressive Multimodal Alignment \citep{zhuang2025math}, automated step-wise pipelines \citep{pan2025enhancing}, GeoLogic for natural-to-formal translation, and in-context learning strategies \citep{xu2024geo}. Benchmarks such as MATHVERSE \citep{mathV} and GeomVerse \citep{kazemi2023geomverse} enable large-scale evaluation. Neuro-symbolic methods combine both paradigms. Alignment improvements \citep{gao2023g}, new datasets \citep{10960701, cho2025geodano}, unified models \citep{cheng2025geouni}, and reinforcement learning approaches \citep{wang2025geometryzero, deng2025openvlthinker} have been explored. Further advances include automated pipelines \citep{huang2025autogeo}, visual augmentation \citep{li2025vision}, and AutoGPS \citep{ping2025autogps}, which leverages ground-truth formalisms but requires costly manual annotations. Overall, symbolic methods are interpretable but rigid, neural models are flexible but error-prone, and neuro-symbolic systems balance the two but often rely on manual inputs. Yet, despite rapid progress in multimodal LLMs, there has been little systematic investigation into whether decomposition into multiple agents actually provides consistent benefits over strong single-agent models in zero-shot geometry problem solving. 

\begin{figure*}[t] 
  \centering
  \includegraphics[width=1\textwidth]{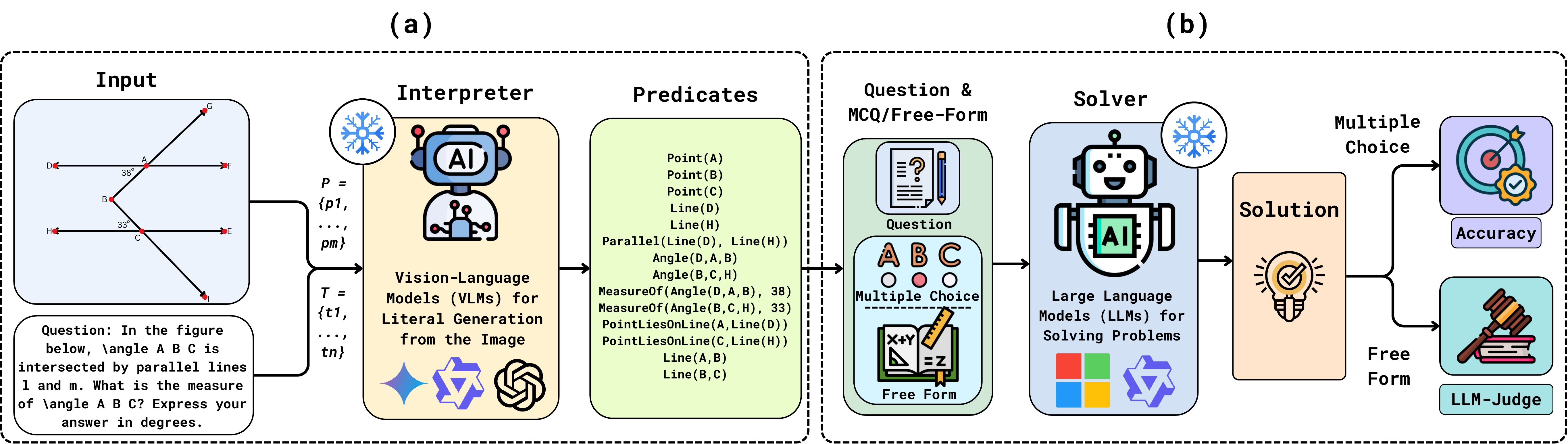} 
  \caption{\textbf{(a)} An Interpreter Agent generates formal predicates from images and questions using VLMs. \textbf{(b)} A Solver Agent then solves the problem using these predicates as LLM input.}
  \label{fig:method}
\end{figure*}

\section{Methodology}\label{method}

In this section, we outline the proposed pipeline for geometry problem solving, shown in Figure~\ref{fig:method}.

\subsection{Problem Formulation}
We study the task of diagram-grounded geometry problem solving, where each problem consists of an image $\text{IMG} = \{\text{img}_1, \text{img}_2, \dots, \text{img}_n\}$ paired with a corresponding question $Q = \{q_1, q_2, \dots, q_n\}$. The objective is to predict the correct solution $\hat{Y}$ for each $(\text{img}_i, q_i)$ pair. This setup allows us to compare different modeling paradigms: \textbf{single-agent} models that directly map from $(\text{img}, q)$ to $\hat{Y}$, and \textbf{multi-agent} pipelines that decompose the task into intermediate steps.

\subsection{Single-Agent}
In the single-agent setting, a single MLLM processes both the geometric image and the textual question end-to-end. Given an input pair $(img, q)$ along with a zero-shot prompt $P$, the model produces a direct prediction:
\begin{equation}
\hat{Y} = MLLM(T_{img}(img), T_{text}([q, P]); W^\star),
\end{equation}
where $T_{img}$ and $T_{text}$ are the respective tokenizers, and $W^\star$ denotes the frozen model parameters.  
This setup does not rely on explicit intermediate representations to solve the problem directly.

\subsection{Multi-Agent}
In the multi-agent setting, we explicitly decompose the task into two stages using two agents: (1) \textbf{Interpreter Agent}: a vision-language model (VLM) that generates symbolic literals that describe the diagram, and (2) \textbf{Solver Agent}: a large language model (LLM) that reasons over these literals and solves the problem. 
The VLM first receives the diagram $img$ and question $q$ along with a zero-shot parsing prompt $P_1$, and autoregressively generates formal literals $\hat{Y}_{vl} = \{l_1, l_2, \dots, l_m\}$:
\begin{equation}
\hat{Y}_{vl} = VL(T_{vl_{img}}(img), T_{vl_{text}}([q, P_1]); W^{vl^\star}).
\end{equation}
These literals capture the geometric relationships present in the diagram.  
Next, the LLM receives $\hat{Y}_{vl}$, the original question $q$, and a problem-solving prompt $P_2$, and produces the final solution:
\begin{equation}
\hat{Y} = LM(T_{lm}([\hat{Y}_{vl}, q, P_2]); W^{lm^\star}).
\end{equation}
This two-stage pipeline allows the VLM to specialize in visual interpretation while the LLM focuses on symbolic reasoning. We experiment with different Interpreter–Solver pairings, including open-source and closed-source models.

\section{Experimental Setup}\label{exp_setup}



\subsection{Datasets}

We evaluate our approaches on four visual math benchmarks: Geometry3K, MathVerse, OlympiadBench, and We-Math. Geometry3K \citep{lu-etal-2021-inter} contains 3{,}001 high-school geometry problems with paired text and diagrams; we follow the official split and report results on the 601 multiple-choice test problems. MathVerse \citep{mathV} has 2{,}612 carefully curated visual math problems with multiple text–diagram configurations; we use the official \texttt{mini-test} subset of 788 problems (436 multiple-choice, 352 free-form). To probe more challenging and less-contaminated settings, we use OlympiadBench \citep{he-etal-2024-olympiadbench}, an Olympiad-level bilingual multimodal benchmark with 8{,}476 math and physics problems and expert step-by-step solutions, and We-Math \citep{qiao-etal-2025-math}, a 6.5K visual math benchmark spanning 67 hierarchical concepts with problem–subproblem decompositions and a four-way diagnostic taxonomy that supports fine-grained analysis of LMMs’ reasoning behavior. 

\subsection{Models}
Our experiments include both open-source and closed-source models to capture the trade-offs between single-agent and multi-agent paradigms. For open-source vision–language models, we use Qwen2.5VL-7B/32B, chosen for their strong zero-shot multimodal reasoning abilities. To make efficient, all open-source models are quantized to 4-bit precision using the \texttt{unsloth} library. For closed-source evaluation, we employ Gemini-2.0-Flash and GPT-4o, the state-of-the-art vision–language models. In the multi-agent pipeline, we experiment with LLMs such as Phi-4 and Qwen3-8B as the solver agent.

\subsection{Evaluation Metrics} 

\paragraph{Multiple-choice tasks.}
We report accuracy as the primary metric. A prediction is considered correct if the model’s output matches one of the provided options. For numerical responses, equivalence is accepted if the predicted value matches the ground truth within the tolerance of the answer choices.  

\paragraph{Free-form tasks.}  
For free-form problems, where answers are not constrained to fixed options, we use accuracy but rely on \textbf{LLM-as-a-judge} using Gemini-2.0-Flash. The judge function $J$ compares the model’s output $A_{llm}$ with the ground truth $A_{gt}$ via a valuation function $v(\cdot)$ that maps responses to numerical form. A response is correct if:
\begin{equation}
J:(A_{llm},A_{gt}) \mapsto (R, [\|v(A_{llm}) - v(A_{gt})\| \le \epsilon]),
\label{eq:judge_metric}
\end{equation}
where $\epsilon$ is a predefined tolerance, $R$ is the reasoning trace, and the binary outcome is 1 for correct and 0 for incorrect. Ambiguous or unverifiable cases are conservatively treated as incorrect.  

\paragraph{Pass@3.}  
Performance for tasks involving MLLMs and LLMs is evaluated using the \texttt{Pass@3} metric, which measures whether a problem is solved in at least one of three independent attempts. 

\section{Results \& Analysis}\label{result}

\subsection{Quantitative Analysis}

\begin{table*}[t]
  \centering
  \tiny
  \begin{adjustbox}{width=\textwidth}
    \begin{tabular}{l| c| cccccc}
      \toprule
      \multicolumn{8}{c}{\textbf{Geometry3K}} \\
      \midrule
      \textbf{Solver} & \textbf{\#Params.} & \multicolumn{3}{c}{\textbf{Multi-Agent}} & \multicolumn{3}{c}{\textbf{Single-Agent}} \\
      \midrule
      \texttt{Qwen-2.5-VL} & 7B  & \multicolumn{3}{c}{\textbf{60.07\%} {\color{Green}(+6.8)}} & \multicolumn{3}{c}{53.24\%} \\
      \texttt{Qwen-2.5-VL} & 32B & \multicolumn{3}{c}{\textbf{72.05\%} {\color{Green}(+3.3)}} & \multicolumn{3}{c}{68.72\%} \\
      \texttt{Gemini-2.0-Flash} & N/A & \multicolumn{3}{c}{83.86\% {\color{red}(–1.3)}} & \multicolumn{3}{c}{\textbf{85.19\%}} \\
      \midrule[\heavyrulewidth]
      
      \multicolumn{8}{c}{\textbf{MathVerse}} \\
      \midrule
      \multirow{2}{*}{\textbf{Solver}} & \multirow{2}{*}{\textbf{\#Params}} & \multicolumn{3}{c}{\textbf{Multi-Agent}} & \multicolumn{3}{c}{\textbf{Single-Agent}} \\
      \cmidrule(lr){3-5} \cmidrule(lr){6-8}
       & & Multiple Choice & Free Form & Overall & Multiple Choice & Free Form & Overall \\
      \midrule
      \texttt{Qwen-2.5-VL} & 7B  & 53.67\% & 36.93\% & 46.19\% {\color{red}(–6.0)} & \textbf{58.94\%} & \textbf{43.75\%} & \textbf{52.16\%} \\
      \texttt{Qwen-2.5-VL} & 32B & \textbf{78.44\%} & \textbf{54.55\%} & \textbf{67.77\%} {\color{Green}(+1.1)} & 76.38\% & 54.55\% & 66.67\% \\
      \texttt{Gemini-2.0-Flash} & N/A & 84.81\% & \textbf{63.48\%} & 74.68\% {\color{red}(-0.45)} & \textbf{86.01\%} & 61.65\% & \textbf{75.13\%} \\
      \midrule[\heavyrulewidth]

      \multicolumn{8}{c}{\textbf{OlympiadBench}} \\
      \midrule
      \textbf{Solver} & \textbf{\#Params} & \multicolumn{3}{c}{\textbf{Multi-Agent}} & \multicolumn{3}{c}{\textbf{Single-Agent}} \\
      \midrule
      \texttt{Qwen-2.5-VL} & 7B  & \multicolumn{3}{c}{\textbf{61.84\%} {\color{Green}(+9.4)}}  & \multicolumn{3}{c}{52.44\%} \\
      \texttt{Qwen-2.5-VL} & 32B & \multicolumn{3}{c}{\textbf{64.56\%} {\color{Green}(+6.67)}} & \multicolumn{3}{c}{57.89\%} \\
      \texttt{Gemini-2.0-Flash} & N/A & \multicolumn{3}{c}{71.31\% {\color{red}(-2.46)}} & \multicolumn{3}{c}{\textbf{73.77\%}} \\
      \midrule[\heavyrulewidth]

      \multicolumn{8}{c}{\textbf{We-Math}} \\
      \midrule
      \textbf{Solver} & \textbf{\#Params} & \multicolumn{3}{c}{\textbf{Multi-Agent}} & \multicolumn{3}{c}{\textbf{Single-Agent}} \\
      \midrule
      \texttt{Qwen-2.5-VL} & 7B  & \multicolumn{3}{c}{\textbf{45.79\%} {\color{Green}(+2.66)}} & \multicolumn{3}{c}{43.13\%} \\
      \texttt{Qwen-2.5-VL} & 32B & \multicolumn{3}{c}{\textbf{59.01\%} {\color{Green}(+4.64)}} & \multicolumn{3}{c}{54.37\%} \\
      \texttt{Gemini-2.0-Flash} & N/A & \multicolumn{3}{c}{\textbf{62.90\%} {\color{Green}(+1.74)}} & \multicolumn{3}{c}{61.16\%} \\
      \bottomrule
    \end{tabular}
  \end{adjustbox}
  \caption{Performance comparison of multi-agent and single-agent approaches on Geometry3K, MathVerse, OlympiadBench, and We-Math. The best score in each row is highlighted in \textbf{bold}. For all multi-agent configurations, the Interpreter agent is fixed to \texttt{Gemini-2.0-Flash}.}
  \label{tab:single_vs_multi}
\end{table*}


Table~\ref{tab:single_vs_multi} provides insight into our central research question: whether multi-agent pipelines outperform single-agent approaches. For open-source models, multi-agent generally improves performance across benchmarks. On Geometry3K, Qwen-2.5-VL (7B) gains +6.8\% (60.07\% vs.\ 53.24\%) and Qwen-2.5-VL (32B) gains +3.3\% (72.05\% vs.\ 68.72\%). On MathVerse, Qwen 32B also benefits (+1.1\%), while the smaller Qwen 7B shows a –6.0\% drop, indicating some sensitivity to model scale and dataset complexity. On OlympiadBench, multi-agent yields sizable improvements for Qwen-2.5-VL 7B (+9.4\%, 61.84\% vs.\ 52.44\%) and 32B (+6.67\%, 64.56\% vs.\ 57.89\%). For Gemini-2.0-Flash, the single-agent configuration remains stronger on Geometry3K, MathVerse, and OlympiadBench (e.g., 85.19\% vs.\ 83.86\% on Geometry3K), but on the newer We-Math benchmark, multi-agent improves performance for all three models: Qwen-2.5-VL 7B (+2.66\%, 45.79\% vs.\ 43.13\%), Qwen-2.5-VL 32B (+4.64\%, 59.01\% vs.\ 54.37\%), and Gemini-2.0-Flash (+1.74\%, 62.90\% vs.\ 61.16\%). Overall, multi-agent decomposition is consistently helpful for open-source models and can also provide modest gains for strong closed-source models on newer visual math benchmarks, but it is not uniformly superior across all architectures and datasets.

\begin{table}[h!]
  \centering
  \tiny
  \begin{adjustbox}{width=0.48\textwidth}
  \begin{tabular}{lc|c}
    \hline
    \textbf{Method} & \textbf{\#Params.} & \multicolumn{1}{c}{\textbf{Accuracy}} \\
    \hline
    \multicolumn{3}{c}{\textbf{Geometry3K}} \\
    \hline
    \verb|Inter-GPS| \cite{lu-etal-2021-inter}        & 406M            & 57.5\% \\
    \verb|GeoDRL| \cite{peng-etal-2023-geodrl}         &  44M            & 68.4\% \\
    \verb|AutoGPS| \cite{ping2025autogps}               & $\approx$200B   & 81.6\% \\
    \verb|Interpreter-Solver-Phi-4 (Ours)|               & 14B-4bit        & 70.05\% \\
    \verb|Interpreter-Solver-Qwen-3 (Ours)|              &  8B-4bit        & 79.53\% \\
    \verb|Interpreter-Solver-Gemini (Ours)|     & N/A   & 83.19\% \\
    \hline
    \multicolumn{3}{c}{\textbf{MathVerse}} \\
    \hline
    \verb|G-LLaVa| \cite{gao2023g}                     & 13B             & 16.6\% \\
    \verb|MathVerse| \cite{mathV}                     &  7B             & 25.9\% \\
    \verb|OpenVLThinker| \cite{deng2025openvlthinker}  &  7B             & 47.9\% \\
    \verb|Interpreter-Solver-Qwen-3 (Ours)|              &  8B-4bit        & 69.67\% \\
    \hline
  \end{tabular}
  \end{adjustbox}
  \caption{Comparison of our multi-agent {\methodname} approach with existing methods.}
  \label{tab:quantitative_results_clean}
\end{table}

Table~\ref{tab:quantitative_results_clean} benchmarks our approach against existing methods on Geometry3K and MathVerse. Consistent with prior literature, stronger models outperform earlier systems such as \verb|Inter-GPS| and \verb|GeoDRL|, but our pipeline establishes a new performance frontier in zero-shot settings. For example, Interpreter–Solver with Gemini-2.0-Flash reaches 83.19\% accuracy on Geometry3K, surpassing \verb|AutoGPS| while using fewer parameters, and Qwen-based variants achieve competitive performance despite being heavily quantized to 4-bit precision. On MathVerse, our Qwen Interpreter–Solver system achieves 69.67\%, representing a substantial gain over prior models such as \verb|OpenVLThinker| (47.9\%) and the MathVerse baseline (25.9\%).  


Taken together, these results highlight a nuanced trade-off. Multi-agent pipelines clearly benefit open-source models by adding structure through explicit intermediate literals. For highly optimized proprietary systems, single-agent reasoning remains stronger on classic benchmarks, with multi-agent offering only modest gains on newer datasets. Thus, agentic decomposition is not universally optimal; its value depends on both the model architecture and the target benchmark. 

\subsection{Predicate Alignment Analysis}
To better understand how the quality of Interpreter-generated literals influences downstream reasoning, we conducted a direct semantic alignment analysis. Since the datasets do not provide gold predicate annotations, we evaluated literal quality by comparing natural-language descriptions derived from two sources: (1) the original diagram and question, and (2) the Interpreter-generated predicates. For each problem, we first generated a reference description from the image and question using \texttt{Gemini-2.5-Flash}, and then generated a second description from the Interpreter's predicates. We embedded both descriptions using OpenAI's \texttt{text-embedding-3-large} model and computed cosine similarity between their embeddings as a proxy for semantic fidelity. As shown in Table~\ref{tab:predicate_alignment}, Gemini-generated predicates achieve the highest average similarity (0.849), followed by GPT-4o mini (0.794) and Qwen-2.5 (0.677). A representative example is provided in Appendix~\ref{app:predicate_example}.

\begin{table}[h!]
\centering
\small
\begin{tabular}{l|c}
\toprule
\textbf{Interpreter} & \textbf{Avg. Cosine Similarity} \\
\midrule
\texttt{Gemini-2.0-Flash} & \textbf{0.849} \\
\texttt{GPT-4o mini}      & 0.794 \\
\texttt{Qwen-2.5}         & 0.677 \\
\bottomrule
\end{tabular}
\caption{Semantic alignment between natural-language descriptions derived from (i) diagram+question and (ii) Interpreter-generated predicates.}
\label{tab:predicate_alignment}
\end{table}

\subsection{Ablation Study}
In table~\ref{tab:ablation_interpreter_solver}, we examine how the choice of Interpreter affects downstream Solver performance. We observe a clear monotonic trend: as the Interpreter becomes stronger, both Solvers (Phi-4 and Qwen-3) improve consistently. When Qwen-2.5-7B is used as the Interpreter, the multi-agent pipeline achieves only 35.77\% and 42.26\% accuracy with Phi-4 and Qwen-3, respectively. Scaling the Interpreter to Qwen-2.5-32B substantially boosts performance, and replacing it with GPT-4o mini yields further gains. The best results are obtained when Gemini serves as the Interpreter, reaching 70.05\% (Phi-4) and 79.53\% (Qwen-3). This ablation confirms that multi-agent effectiveness is tightly coupled to the quality of the Interpreter’s literals: weak Interpreters bottleneck the pipeline, whereas strong ones unlock the full potential of the Solver.

\begin{table}[h!]
  \centering
  \tiny
  \begin{adjustbox}{width=0.40\textwidth}
  \begin{tabular}{cc|cc}
    \hline
    \textbf{Interpreter} & \textbf{\#Params.}
    & \multicolumn{2}{c}{\textbf{Solver}} \\
    \cline{3-4}
    &
    & \verb|Phi-4|
    & \verb|Qwen 3| \\
    \hline
    \verb|Qwen-2.5|
    & 7B
    & 35.77\%
    & 42.26\% \\

    \verb|Qwen-2.5|
    & 32B
    &  56.74\%
    & 61.23\% \\

    \verb|GPT-4o mini|
    & $\approx$8B
    & 58.24\%
    & 63.23\% \\

    \verb|Gemini|
    & N/A
    & 70.05\%
    & 79.53\%\\
    \hline
  \end{tabular}
  \end{adjustbox}
  \caption{Comparison of the accuracy of different {\methodname} settings.}
  \label{tab:ablation_interpreter_solver}
\end{table}

\section{Discussion}\label{discussion}

Our analysis highlights a nuanced trade-off between single-agent and multi-agent pipelines for geometry and visual math problem solving. Multi-agent decomposition, which separates perception and reasoning, tends to help open-source models, especially at medium scale and on harder benchmarks. For example, Qwen-2.5-VL-32B gains on Geometry3K, OlympiadBench, and We-Math when guided by the Interpreter’s literals, suggesting that explicit structure can stabilize reasoning and reduce ambiguity in multi-step configurations. In qualitative cases, we observe that the Interpreter’s predicates often prevent Qwen-2.5-VL-32B from drifting into inconsistent chains of thought by anchoring it to a small set of geometric relations.

However, multi-agent design is not universally beneficial, and its effectiveness depends strongly on model capacity and literal quality. For smaller models such as Qwen-2.5-VL-7B on MathVerse, decomposition can introduce noisy or over-constraining predicates, leading to measurable drops in accuracy compared to single-agent mode. For Gemini-2.0-Flash, which already couples perception and reasoning tightly, single-agent pipelines remain stronger on classic benchmarks, with multi-agent only yielding modest gains on We-Math. In several error cases, overly detailed or partially incorrect literals caused Gemini-2.0-Flash and Qwen-2.5-VL-32B to overfit local constraints (e.g., misusing an angle label or misreading a ratio), underscoring that agentic decomposition is most effective when predicates are compact, accurate, and aligned with the model’s internal reasoning style.





\section{Conclusion}\label{conclusion}

We presented a systematic comparison of single-agent and multi-agent pipelines for diagram-grounded geometry and visual math problem solving. Across four benchmarks, multi-agent decomposition consistently benefits open-source models, especially at medium scale, while strong proprietary systems often remain stronger in single-agent mode, with multi-agent offering only modest gains on newer datasets. These results show that agentic decomposition is helpful but not universally optimal, with its value depending on model architecture and benchmark characteristics. A natural next step is to develop adaptive strategies that select between single-agent and multi-agent configurations based on model capacity and task demands.

\clearpage
\newpage

\section*{Limitations}
While our study offers new insights into the trade-offs between single-agent and multi-agent pipelines for geometrical problem-solving, it has several limitations. First, our analysis was conducted exclusively in a zero-shot setting, with all model parameters frozen. This prevents us from exploring whether fine-tuning could alter the relative advantage of single-agent versus multi-agent approaches. Second, our methodology was restricted to a fixed prompting setup. Adaptive or iterative prompting strategies could potentially change how each paradigm performs by refining reasoning step by step. Thirdly, our study examined a limited set of models. Including more model families and scales would provide a broader perspective on when decomposition helps or hinders. Fourth, the quality of generated literals played a central role in multi-agent outcomes, but we did not systematically evaluate alternative extraction strategies. Finally, our experiments were constrained by resources, requiring 4-bit quantized models via the Unsloth library. Thus, our findings based on open-source models may not fully reflect the behavior of larger, full-precision systems or state-of-the-art proprietary models.


\bibliography{custom}
\clearpage
\newpage

\onecolumn
\appendix
\section{Appendix}

\subsection{Predicate Alignment Example}
\label{app:predicate_example}
We validate the semantic fidelity of our symbolic predicates against the original visual data. Using Problem 2405 as a representative example, we compare the 'Description from Diagram' against a synthetic 'Description from Gemini Predicates.' The high cosine similarity ($\approx 0.856$) confirms that the symbolic predicates successfully capture the essential geometric relationships and spatial configurations present in the image, effectively bridging the modality gap.

\begin{tcolorbox}[colback=white!5,colframe=red!50,title={Problem Number 2405 from Geometry3K Dataset}]
\small
\begin{center}
\includegraphics[width=0.35\textwidth]{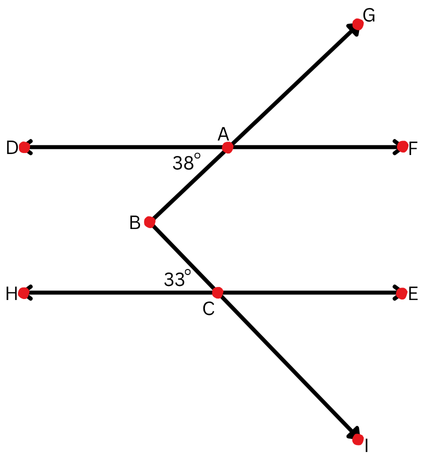}
\end{center}

\textbf{Question:} \\
In the figure below, $\angle$ A B C is intersected by parallel lines l and m. What is the measure of $\angle$ A B C? Express your answer in degrees. \\[0.6em]

\textbf{Predicate:} \\
\texttt{Point(A)} \\
\texttt{Point(B)} \\
\texttt{Point(C)} \\
\texttt{Line(D)} \\
\texttt{Line(H)} \\
\texttt{Parallel(Line(D), Line(H))} \\
\texttt{Angle(D,A,B)} \\
\texttt{Angle(B,C,H)} \\
\texttt{MeasureOf(Angle(D,A,B), 38)} \\
\texttt{MeasureOf(Angle(B,C,H), 33)} \\
\texttt{PointLiesOnLine(A,Line(D))} \\
\texttt{PointLiesOnLine(C,Line(H))} \\
\texttt{Line(A,B)} \\
\texttt{Line(B,C)} \\[0.6em]

\textbf{Description from Diagram + Question:} \\
A geometry diagram shows angle ABC formed by two rays BA and BC that cross two parallel lines $l$ and $m$. Point A is on the upper parallel line $l$ and point C is on the lower parallel line $m$. On line $l$, to the left of A, there is a point D so that ray AD lies along line $l$. On line $m$, to the right of C, there is a point H so that ray CH lies along line $m$. Segment AB slants up and to the right, and segment CB slants up and to the left, so that vertex B is between the two parallel lines. The acute angle between line $l$ and segment BA at A is marked as $38^\circ$. The acute angle between line $m$ and segment CB at C is marked as $33^\circ$. \\[0.6em]
\textbf{Description from Gemini Predicates:} \\
The configuration has three points A, B, and C and two straight lines D and H that are parallel to each other. Point A lies on line D and point C lies on line H, while B is between them so that segments AB and CB meet at B to form the angle of interest. There is an angle DAB at point A between the ray along line D and segment AB whose measure is $38^\circ$. There is another angle BCA at point C between segments CB and CA whose measure is $33^\circ$. Segments AB and BC thus connect the upper line D to the lower line H, acting as a broken transversal across the two parallel lines, and the given $38^\circ$ and $33^\circ$ angles can be related to the angle at B. \\[0.6em]

\textbf{Cosine similarity:} $\approx 0.856$
\end{tcolorbox}




\subsection{Prompt Templates}
\subsubsection{Predicate Generation Prompt}
To extract accurate symbolic representations from geometric diagrams without task-specific fine-tuning, we employ a structured prompting strategy. By instructing the Visual Language Model (VLM) to adopt the persona of an expert mathematician and requiring a "Step-by-Step Analysis" prior to symbolic generation, we mitigate hallucination and ensure the model explicitly grounds its logic in observed visual features ranging from basic shape identification to complex constraints.

\begin{tcolorbox}[
  colback=blue!5!white,
  colframe=blue!30!black,
  boxrule=0.5pt,
  arc=3pt,
  left=5pt, right=5pt, top=5pt, bottom=5pt,
  enhanced,
  breakable,
  listing only,
  listing options={basicstyle=\ttfamily\small,breaklines=true}
]
\begin{verbatim}
You are an expert AI mathematician specializing in geometry. Your task is to analyze 
the geometric figure in the provided image and generate accurate geometric predicates
(literals) that represent ALL the relationships, measurements, and properties shown 
in the diagram.

GEOMETRY PROBLEM IMAGE:
The image shows a geometric figure with various shapes, lines, angles, and 
measurements.
Analyze this image carefully to understand all geometric relationships and 
constraints.

Question: [Given Question]

YOUR TASK:
1. First, provide step-by-step reasoning showing your analysis process
2. Then, generate geometric predicates based on your analysis using the Guidelines 
below

STEP-BY-STEP ANALYSIS (Required):
Please follow this format for your reasoning:

1. COMPREHENSIVE IMAGE ANALYSIS
   - Identify ALL geometric shapes (circles, triangles, quadrilaterals, etc.)
   - List ALL points, lines, and their labels or names
   - Note ALL visible measurements, angles, and numerical values
   - Identify ALL special markings (right angle symbols, parallel marks,
     congruent marks, equal marks, etc.)
   - Look for implied constructions (perpendiculars, bisectors, tangents,
     chords, radii, etc.)

2. CIRCLE-SPECIFIC ANALYSIS (If circles are present)
   - Identify the center and all points on the circle
   - Determine which lines are radii, chords, diameters, or tangents
   - Look for inscribed angles, central angles, and arc relationships
   - Check for perpendicular relationships involving radii and chords
   - Identify any equal radius relationships

3. ANGLE AND PERPENDICULARITY ANALYSIS
   - Examine ALL angles shown in the diagram, both marked and unmarked
   - Look for right-angle indicators or perpendicular relationships
   - Check for angle bisectors or special angle relationships
   - Identify complementary, supplementary, or vertical angles
   - Look for inscribed angles and their corresponding arcs

4. CONGRUENCE AND EQUALITY ANALYSIS
   - Identify ALL equal lengths, angles, or shapes
   - Check for congruent triangles or similar figures
   - Look for equal radii in circles
   - Identify parallel lines or equal distances

5. INTERSECTION AND POSITIONING ANALYSIS
   - Determine where lines intersect and at what points
   - Check if points lie on specific lines or circles
   - Identify midpoints, centroids, or other special points
   - Look for points that divide segments in specific ratios

6. CONSTRAINT AND RELATIONSHIP SYNTHESIS
   - Combine observations to identify implicit relationships
   - Look for theorem applications (Pythagorean, inscribed angle, etc.)
   - Identify geometric constructions that create specific relationships
   - Check for properties that follow from the given constraints

QUESTION-DRIVEN COMPLETENESS CHECK
1. Ensure all information needed to solve the problem is captured
2. Verify that key relationships for the solution are represented
3. Double-check that no critical geometric properties are missed
4. Confirm that the predicates will provide sufficient information for 
problem-solving

CRITICAL ANALYSIS GUIDELINES:
- LOOK FOR HIDDEN RELATIONSHIPS:
  Many geometric problems have implicit perpendicular relationships,
  equal lengths, or special angle properties that are not explicitly
  marked but are crucial for solving.

- CIRCLE GEOMETRY FOCUS:
  If the diagram contains circles, pay special attention to:
  * Which points lie on the circle versus inside or outside
  * Perpendicular relationships between radii and chords
  * Equal radius lengths
  * Inscribed versus central angles
  * Tangent-radius perpendicularity

- CONSTRUCTION INDICATORS:
  Look for:
  * Lines that appear to be perpendicular even without explicit markings
  * Points that appear to be midpoints or special positions
  * Equal lengths suggested by visual symmetry
  * Angle relationships implied by the construction

GUIDELINES:
Follow these predicates to represent diagram literals.

GEOMETRIC SHAPES:
- Point: Point(A), Point()
- Line: Line(A,B), Line(m), Line()
- Angle: Angle(A,B,C), Angle(A), Angle(1), Angle()
- Triangle: Triangle(A,B,C), Triangle(), Triangle(1,2,3)
- Quadrilateral: Quadrilateral(A,B,C,D), Quadrilateral()
- Parallelogram: Parallelogram(A,B,C,D), Parallelogram(1), Parallelogram()
- Square: Square(A,B,C,D), Square(1), Square()
- Rectangle: Rectangle(A,B,C,D), Rectangle(1), Rectangle()
- Rhombus: Rhombus(A,B,C,D), Rhombus(1), Rhombus()
- Trapezoid: Trapezoid(A,B,C,D), Trapezoid(1), Trapezoid()
- Kite: Kite(A,B,C,D), Kite(1), Kite()
- Polygon: Polygon()
- Pentagon: Pentagon(A,B,C,D,E), Pentagon()
- Hexagon: Hexagon(A,B,C,D,E,F), Hexagon()
- Heptagon: Heptagon(A,B,C,D,E,F,G), Heptagon()
- Octagon: Octagon(A,B,C,D,E,F,G,H), Octagon()
- Circle: Circle(A), Circle(1), Circle()
- Arc: Arc(A,B), Arc(A,B,C), Arc()
- Sector: Sector(O,A,B), Sector()
- Shape: Shape()   // For unknown shapes or regions

UNARY GEOMETRIC ATTRIBUTES:
- RightAngle: RightAngle(Angle())
- Right: Right(Triangle())   // Right triangle
- Isosceles: Isosceles(Polygon())   // Isosceles polygon
- Equilateral: Equilateral(Polygon())   // Equilateral polygon
- Regular: Regular(Polygon())
- Red: Red(Shape())
- Blue: Blue(Shape())
- Green: Green(Shape())
- Shaded: Shaded(Shape())

GEOMETRIC ATTRIBUTES:
- AreaOf: AreaOf(A)
- PerimeterOf: PerimeterOf(A)   // Perimeter of polygon A
- RadiusOf: RadiusOf(A)
- DiameterOf: DiameterOf(A)
- CircumferenceOf: CircumferenceOf(A)   // Perimeter of circle A
- AltitudeOf: AltitudeOf(A)   // Altitude of polygon A
- HypotenuseOf: HypotenuseOf(A)   // Hypotenuse of triangle A
- SideOf: SideOf(A)   // Side of square A
- WidthOf: WidthOf(A)   // Width of quadrilateral A
- HeightOf: HeightOf(A)   // Height of quadrilateral A
- LegOf: LegOf(A)   // Leg of trapezoid A
- BaseOf: BaseOf(A)   // Base of polygon A
- MedianOf: MedianOf(A)   // Median of polygon A
- IntersectionOf: IntersectionOf(A,B)   // Intersection of shapes A and B
- MeasureOf: MeasureOf(A)   // Measure of angle A
- LengthOf: LengthOf(A)   // Length of line A
- ScaleFactorOf: ScaleFactorOf(A,B)   // Scale factor of shape A to shape B

BINARY GEOMETRIC RELATIONS:
- PointLiesOnLine: PointLiesOnLine(Point(), Line(1,2))
- PointLiesOnCircle: PointLiesOnCircle(Point(), Circle())
- Parallel: Parallel(Line(), Line())
- Perpendicular: Perpendicular(Line(), Line())
- IntersectAt: IntersectAt(Line(), Line(), Line(), Point())
- BisectsAngle: BisectsAngle(Line(), Angle())
- Congruent: Congruent(Polygon(), Polygon())
- Similar: Similar(Polygon(), Polygon())
- Tangent: Tangent(Line(), Circle())
- Secant: Secant(Line(), Circle())
- CircumscribedTo: CircumscribedTo(Shape(), Shape())
- InscribedIn: InscribedIn(Shape(), Shape())

A-IsXOf-B GEOMETRIC RELATIONS:
- IsMidpointOf: IsMidpointOf(Point(), Line())
  // Point A is midpoint of line B
- IsCentroidOf: IsCentroidOf(Point(), Shape())
  // Point A is centroid of shape B
- IsIncenterOf: IsIncenterOf(Point(), Shape())
  // Point A is incenter of shape B
- IsRadiusOf: IsRadiusOf(Line(), Circle())
  // Line A is radius of circle B
- IsDiameterOf: IsDiameterOf(Line(), Circle())
  // Line A is diameter of circle B
- IsMidsegmentOf: IsMidsegmentOf(Line(), Triangle())
  // Line A is midsegment of triangle B
- IsChordOf: IsChordOf(Line(), Circle())
  // Line A is chord of circle B
- IsSideOf: IsSideOf(Line(), Polygon())
  // Line A is side of polygon B
- IsHypotenuseOf: IsHypotenuseOf(Line(), Triangle())
  // Line A is hypotenuse of triangle B
- IsPerpendicularBisectorOf: IsPerpendicularBisectorOf(Line(), Triangle())
  // Line A is perpendicular bisector of triangle B
- IsAltitudeOf: IsAltitudeOf(Line(), Triangle())
  // Line A is altitude of triangle B
- IsMedianOf: IsMedianOf(Line(), Quadrilateral())
  // Line A is median of quadrilateral B
- IsBaseOf: IsBaseOf(Line(), Quadrilateral())
  // Line A is base of quadrilateral B
- IsDiagonalOf: IsDiagonalOf(Line(), Quadrilateral())
  // Line A is diagonal of quadrilateral B
- IsLegOf: IsLegOf(Line(), Trapezoid())
  // Line A is leg of trapezoid B

NUMERICAL ATTRIBUTES AND RELATIONS:
- SinOf: SinOf(Var)
- CosOf: CosOf(Var)
- TanOf: TanOf(Var)
- CotOf: CotOf(Var)
- HalfOf: HalfOf(Var)
- SquareOf: SquareOf(Var)
- SqrtOf: SqrtOf(Var)
- RatioOf: RatioOf(Var), RatioOf(Var1, Var2)
- SumOf: SumOf(Var1, Var2, ...)
- AverageOf: AverageOf(Var1, Var2, ...)
- Add: Add(Var1, Var2, ...)
- Mul: Mul(Var1, Var2, ...)
- Sub: Sub(Var1, Var2, ...)
- Div: Div(Var1, Var2, ...)
- Pow: Pow(Var1, Var2)
- Equals: Equals(Var1, Var2)
- UseTheorem: UseTheorem(A_B_C)

VARIABLE NAMING CONVENTIONS:
- Use capital letters for points: A, B, C, D, etc.
- Use lowercase letters for lines when not defined by points: m, n, l, etc.
- Use numbers for unnamed shapes: 1, 2, 3, etc.
- Use $ for generic variables: $, $1, $2, etc.
- Use descriptive names when appropriate: base, height, radius, etc.

CRITICAL INSTRUCTIONS:
1. BE EXTREMELY THOROUGH
   Missing relationships are the main cause of poor problem-solving performance

2. LOOK BEYOND THE OBVIOUS
   Many critical relationships are implied, not explicitly marked

3. Carefully examine the geometric figure in the image

4. Identify all points, lines, angles, shapes, and measurements shown

5. MAKE EACH PREDICATE AS ATOMIC AS POSSIBLE
   - Decompose any complex or compound relationship into the simplest,
     individual geometric statements
   - For example, replace a single perpendicular statement with
     simpler angle-equals-90-degree or vector-based predicates

INSTRUCTIONS FOR PREDICATE GENERATION:

1. Generate predicates that represent:
   - All geometric shapes present
   - All given measurements and their relationships
   - All geometric properties and constraints, including implied ones
   - ALL relationships between different elements
   - All perpendicular relationships, both marked and implied
   - All equal lengths and angles, both marked and implied

2. Always provide the step-by-step reasoning first

3. Then provide the predicates section with a clear section header

4. Follow the Guidelines above
   These predicates are crucial for representing diagram literals

5. Each predicate must be on a separate line

6. Do not include quotation marks, extra symbols, or explanatory text in predicates

7. Only output predicates in the exact format:
   PredicateName(arguments)

8. IMPORTANT:
   Do NOT include Find(...) predicates or any question-related predicates

9. Include only the given information, constraints, and geometric relationships 
   visible in the diagram

10. Represent all visible geometric relationships, not derived solutions

11. The predicates should provide sufficient information for another system to solve
    the problem, but not the solution itself

12. COMPLETENESS IS KEY
    It is better to include extra relationships than to miss critical ones.
\end{verbatim}
\end{tcolorbox}

\subsubsection{Multiple Choice Geometry Problem Solving Prompt}
The prompt instructs the model to perform systematic geometric reasoning using a multi-step analytical framework. The following standardized prompt was used to decompose geometric predicates and apply mathematical theorems to solve multiple-choice problems.

\begin{tcolorbox}[
  colback=blue!5!white,
  colframe=blue!30!black,
  boxrule=0.5pt,
  arc=3pt,
  left=5pt, right=5pt, top=5pt, bottom=5pt,
  enhanced,
  breakable,
  listing only,
  listing options={basicstyle=\ttfamily\small,breaklines=true}
]
\begin{verbatim}
You are an expert AI mathematician specializing in geometry. Your task is to solve 
the following geometric problem using the provided predicates through systematic 
reasoning and theorem application.

Question:
Predicates: [Given Predicate]
Question: [Given Question]
Choices: [Given Choices]


YOUR TASK:
Provide a complete step-by-step solution following the structured approach below,
then select the correct answer choice.

STEP-BY-STEP SOLUTION PROCESS

STEP 1: PREDICATE ANALYSIS AND SETUP
- Parse and categorize the given predicates into:
  * Geometric shapes (points, lines, circles, triangles, etc.)
  * Measurements and equalities (lengths, angles, areas)
  * Relationships (perpendicular, parallel, congruent, etc.)
  * Positioning (points on lines or circles, intersections, etc.)
- Identify what specific value or measurement the question is asking for.
- Note any special geometric constructions or theorems that might apply.

STEP 2: CONSTRAINT SYNTHESIS
- Combine related predicates to understand the complete geometric picture.
- Identify key relationships that will be useful for solving.
- Look for:
  * Equal lengths or angles that can be substituted
  * Perpendicular relationships that create right triangles
  * Circle properties (radii, chords, central or inscribed angles)
  * Congruent or similar triangles
  * Theorem applications (Pythagorean, inscribed angle, etc.)

STEP 3: SOLUTION STRATEGY
- Based on the predicates and question, determine the most direct solution path.
- Identify which geometric theorems, properties, or formulas to apply.
- Plan the sequence of logical steps needed to reach the answer.

STEP 4: MATHEMATICAL DERIVATION
- Execute your solution strategy step by step.
- Show all calculations clearly.
- Apply geometric theorems and properties systematically.
- Use the relationships established in the predicates.
- Substitute known values and solve for unknowns.

STEP 5: VERIFICATION AND ANSWER SELECTION
- Verify the calculated result makes geometric sense.
- Compare the result with the provided answer choices.
- Select the choice that best matches the calculated answer.
- If no exact match exists, select the closest reasonable option.

GEOMETRIC REASONING GUIDANCE
- Consider all relevant geometric theorems and properties.
- Apply circle, triangle, quadrilateral, and angle theorems as appropriate.
- Look for relationships between shapes, measurements, and positions.
- Use both basic and advanced geometric principles as needed.

PREDICATE USAGE GUIDANCE
- Interpret predicates based on their geometric meaning and context.
- Combine multiple predicates to understand complex relationships.
- Consider both direct and derived information from predicate combinations.

CRITICAL INSTRUCTIONS:
1. USE THE PREDICATES SYSTEMATICALLY
   Every predicate provides important information
2. APPLY RELEVANT GEOMETRIC KNOWLEDGE
   Use any geometric theorems, properties, or principles that help solve 
   the problem
3. REASON FLEXIBLY
   Adapt your approach based on the specific problem and predicates
4. SHOW ALL WORK
   Make your reasoning clear and mathematical
5. BE PRECISE
   Use exact values when possible, approximate only when necessary
\end{verbatim}
\end{tcolorbox}

\subsubsection{Free-Form Geometry Problem Solving Prompt}
The prompt instructs the model to solve free-form geometric word problems through a multi-step analytical and derivation process. The following standardized prompt was used to guide the model from initial predicate categorization to the final verification of results.

\begin{tcolorbox}[
  colback=blue!5!white,
  colframe=blue!30!black,
  boxrule=0.5pt,
  arc=3pt,
  left=5pt, right=5pt, top=5pt, bottom=5pt,
  enhanced,
  breakable,
  listing only,
  listing options={basicstyle=\ttfamily\small,breaklines=true}
]
\begin{verbatim}
You are an expert AI mathematician specializing in geometry. Your task is to solve 
the following geometric problem using the provided predicates through systematic
reasoning and theorem application.

Question:
Predicates: [Given Predicate]
Question: [Given Question]
Choices: [Given Choices]


YOUR TASK:
Provide a complete step-by-step solution following the structured approach below,
then provide your final answer in proper mathematical format.

STEP-BY-STEP SOLUTION PROCESS:

STEP 1: PREDICATE ANALYSIS AND SETUP
- Parse and categorize the given predicates into:
  * Geometric shapes (points, lines, circles, triangles, etc.)
  * Measurements and equalities (lengths, angles, areas)
  * Relationships (perpendicular, parallel, congruent, etc.)
  * Positioning (points on lines or circles, intersections, etc.)
- Identify what specific value or measurement the question is asking for
- Note any special geometric constructions or theorems that might apply

STEP 2: CONSTRAINT SYNTHESIS
- Combine related predicates to understand the complete geometric picture
- Identify key relationships that will be useful for solving
- Look for:
  * Equal lengths or angles that can be substituted
  * Perpendicular relationships that create right triangles
  * Circle properties (radii, chords, central or inscribed angles)
  * Congruent or similar triangles
  * Theorem applications (Pythagorean, inscribed angle, etc.)

STEP 3: SOLUTION STRATEGY
- Based on the predicates and question, determine the most direct solution path
- Identify which geometric theorems, properties, or formulas to apply
- Plan the sequence of logical steps needed to reach the answer

STEP 4: MATHEMATICAL DERIVATION
- Execute your solution strategy step by step
- Show all calculations clearly
- Apply geometric theorems and properties systematically
- Use the relationships established in the predicates
- Substitute known values and solve for unknowns

STEP 5: VERIFICATION AND FINAL ANSWER
- Verify your calculated result makes geometric sense
- Express your final answer in proper mathematical format
- Ensure units are included when applicable
- Round to appropriate precision when necessary

GEOMETRIC REASONING GUIDANCE:
- Consider all relevant geometric theorems and properties
- Apply circle, triangle, quadrilateral, and angle theorems as appropriate
- Look for relationships between shapes, measurements, and positions
- Use both basic and advanced geometric principles as needed

PREDICATE USAGE GUIDANCE:
- Interpret predicates based on their geometric meaning and context
- Combine multiple predicates to understand complex relationships
- Consider both direct and derived information from predicate combinations

CRITICAL INSTRUCTIONS:
1. USE THE PREDICATES SYSTEMATICALLY
   Every predicate provides important information
2. APPLY RELEVANT GEOMETRIC KNOWLEDGE
   Use any geometric theorems, properties, or principles that help solve the 
   problem
3. REASON FLEXIBLY
   Adapt your approach based on the specific problem and predicates
4. SHOW ALL WORK
   Make your reasoning clear and mathematical
5. BE PRECISE
   Use exact values when possible, approximate only when necessary
\end{verbatim}
\end{tcolorbox}



\subsection{Qualitative Analysis of Model Reasoning and Error Patterns}
The following examples illustrate model outputs across geometric reasoning tasks. These traces provide a qualitative comparison between multiple-choice and free-form examples \ref{tab:ex_328}, highlighting behavioral phenomena such as recursive self-doubt \ref{tab:RQ_2}, reasoning loops \ref{tab:RQ_3}, and erroneous reassessment steps \ref{tab:RQ_4}.
\begin{table*}[!htbp]
  \centering
  \begin{adjustbox}{width=\textwidth}
  \begin{tcolorbox}[
    colback=yellow!5!white,        
    colframe=black!50!black,      
    colbacktitle=yellow!20!white,  
    coltitle=black,              
    fonttitle=\bfseries,         
    title=Reasoning comparison for MathVerse example 328 with and without choices.,
    halign title=center,
    fontupper=\small,            
    enhanced,                    
    boxrule=0.5pt                
  ]
  \begin{tabular}{@{}p{0.48\linewidth}||p{0.48\linewidth}@{}}
    \textbf{Wrong Reasoning} & \textbf{Right Reasoning} \\

    \hline
    \\
    
    \textbf{PREDICATES:} 
    Point(A),
    Point(B),
    Point(C),
    Line(A,B),
    Line(A,C),
    Line(B,C),
    Triangle(A,B,C),
    RightAngle(Angle(A,B,C)),
    LengthOf(Line(A,B),17.6),
    LengthOf(Line(A,C),d),
    \colorbox{red!30}{MeasureOf(Angle(B,A,C),52)}, 
    Perpendicular(Line(A,B),Line(B,C))
    
    \textbf{QUESTION:}
    Three television presenters are practising their navigation skills before heading off on an expedition to a remote location. Amelia at point B is positioned 17.6 metres south of Ned at point A. \colorbox{red!30}{Bart at point C is due east of Amelia and on a bearing} \colorbox{red!30}{of S $38^{\circ}$ E from Ned}. If Ned and Bart are d metres apart, find d to one decimal place.
    Okay, let me try to figure out this problem. So, we have three points: A, B, and C. Let me start by visualizing the setup. 
    First, the problem says Amelia is at point B, and she's positioned 17.6 meters south of Ned at point A. So, if I imagine a coordinate system where north is up, then point A is north of point B. Wait, but directions can sometimes be confusing. Let me clarify: "south" means that B is directly south of A. So, if I consider A at some coordinate, then B would be straight down (south) from A. $\ldots$ $\ldots$ $\ldots$ 
    
    But the problem says that MeasureOf(Angle(B,A,C),52). This implies that $\alpha$ = 52 degrees. 
    \colorbox{Green!30}{This contradiction suggests that either the problem} has conflicting information, or I have misinterpreted the bearing.
    Wait, could the bearing be referring to the angle between AC and the north direction? Let me think. If the bearing is S $38^{\circ}$ E, then the angle between AC and south is 38 degrees. But the angle at A between AB (south) and AC is 38 degrees. So, this would mean $\alpha$ = 38 degrees. But the problem says it's 52 degrees. Unless... there's a miscalculation in the problem's given information. Alternatively, maybe the angle mentioned in the problem is not the angle at A, but at C? Wait, the problem says: MeasureOf(Angle(B,A,C),52). This is angle at A between B and C, so it's angle BAC, which should be 38 degrees according to the bearing. But the problem says 52. 
    \colorbox{red!30}{So, using the right triangle} with $\angle A = 52^\circ$ and $AB = 17.6\,$m, we have $AC = \frac{AB}{\cos(52^\circ)}$; since $\cos(52^\circ)\approx0.6157$, it follows that \colorbox{red!30}{$d = \frac{17.6}{0.6157}\approx28.6\,$m}. Although the problem also gives a \colorbox{Green!30}{bearing of S~$38^\circ$~E} (which would yield a different result), the explicit angle \colorbox{red!30}{at $A$ of $52^\circ$ takes precedence.} Therefore, despite the apparent contradiction, the correct value of $d$ is $\frac{17.6}{\cos(52^\circ)}$. Rounded to one decimal place, it would be $28.6$. But wait, let me check with the bearing. If the bearing is S~$38^\circ$~E, then the angle at $A$ is $38^\circ$, so $d = \frac{17.6}{\cos(38^\circ)} \approx \frac{17.6}{0.7880} \approx 22.34$, which rounds to $22.3$. But the problem states the angle at $A$ is $52^\circ$. Since the problem provides both the angle and the bearing, and they seem conflicting, but the predicates include $\mathrm{MeasureOf}(Angle(B,A,C))=52^\circ$, which is the angle at $A$, I must prioritize that. Thus, the answer \colorbox{red!30}{is approximately $28.6\,$m.}

    & 
    
    \textbf{PREDICATES:} 
    Point(A),
    Point(B),
    Point(C),
    Line(A,B),
    Line(A,C),
    Line(B,C),
    Triangle(A,B,C),
    RightAngle(Angle(A,B,C)),
    LengthOf(Line(A,B),17.6),
    LengthOf(Line(A,C),d),
    \colorbox{red!30}{MeasureOf(Angle(B,A,C),52)}, 
    Perpendicular(Line(A,B),Line(B,C))
    
    \textbf{QUESTION:}
    Three television presenters are practising their navigation skills before heading off on an expedition to a remote location. Amelia at point B is positioned 17.6 metres south of Ned at point A. \colorbox{red!30}{Bart at point C is due east of Amelia and on a bearing} \colorbox{red!30}{of S $38^{\circ}$ E from Ned}. If Ned and Bart are d metres apart, find d to one decimal place.
    
    \textbf{}{CHOICES:}
    A. 22.3
    B. 44.5
    C. 20.4
    D. 50

    Okay, let's tackle this problem step by step. So, we have three points: A, B, and C. From the given predicates, I need to figure out the distance between Ned (at point A) and Bart (at point C), which is denoted as 'd'.
    
    $\ldots$
    
    $\ldots$
    
    $\ldots$

    \colorbox{Green!30}{Therefore, there is a contradiction here.} Hence, my assumption must be wrong somewhere. This inconsistency indicates that I've misunderstood either the angle given or the bearing. Let me revisit the problem statement. In triangle $ABC$, $\angle A = 52^\circ$. So, the angle between $BA$ (south) and $AC$ is $52^\circ$. But the bearing says it’s $38^\circ$. \colorbox{Green!30}{So, why is there a discrepancy?} Ah! Wait, maybe I mixed up the angle. The angle at $A$ is between $BA$ and $CA$. $BA$ is south, and $CA$ is going to point $C$. The angle between $BA$ and $CA$ is $52^\circ$, which is the angle between south and $AC$. But the bearing from $A$ to $C$ is S~$38^\circ$~E, which suggests that this angle is $38^\circ$. So, this implies that the given angle of $52^\circ$ conflicts with the bearing of $38^\circ$. This inconsistency suggests that \colorbox{Green!30}{my initial assumptions are incorrect.}

    $\ldots$
    
    $\ldots$
    
    $\ldots$

    \colorbox{Green!30}{So, $d \approx 17.6 / 0.7880 \approx 17.6 / 0.7880 \approx 22.34\,$m.} Looking at the answer choices, option A is $22.3$, which is close. However, given that the only way to get a value matching the answer choices is to take the bearing angle as $38^\circ$, leading to $d \approx 22.3$, which is option A, and assuming that the angle at $A$ is $38^\circ$, even though the problem says it’s $52^\circ$, there might be a typo in the problem. Or perhaps I misread the angle. \colorbox{Green!30}{Thus, the correct answer is A.}

  \end{tabular}
  \end{tcolorbox}
  \end{adjustbox}
  \caption{Comparison of reasoning with and without multiple-choice options for example number 328 from the MathVerse dataset. We have highlighted in \colorbox{red!30}{red} where the Solver's (Qwen) reasoning detected ambiguity and how the influence of the given choices, in \colorbox{Green!30}{green}, corrected.}
  \label{tab:ex_328}
\end{table*}


\begin{table*}[!htbp]
  \centering
  \begin{adjustbox}{width=\textwidth}
  \begin{tcolorbox}[
    colback=yellow!5!white,        
    colframe=black!50!black,      
    colbacktitle=yellow!20!white,  
    coltitle=black,              
    fonttitle=\bfseries,        
    title=Illustration of the model’s recursive self-doubt,
    halign title=center,
    fontupper=\small            
  ]

\textbf{PREDICATES:}
Rhombus(A,B,C,D)
Point(A)
Point(B)
Point(C)
Point(D)
Equals(AreaOf(Rhombus(A,B,C,D)), 36)
Equals(LengthOf(Line(A,D)), 8)
Shape(1)
Equals(LengthOf(AltitudeOf(Shape(1))), 4)
Similar(Rhombus(A,B,C,D), Shape(1))
IsAltitudeOf(Line(A,D), Rhombus(A,B,C,D))
Green(Shape(1))
Blue(Rhombus(A,B,C,D))

\textbf{QUESTION:}
For the pair of similar figures, find the area of the Green figure.

\textbf{CHOICES:}
A. 9
B. 18
C. 72
D. 144

Okay, let's tackle this geometry problem step by step. First, I need to parse all the given information and constraints. Let me start by listing out everything provided.

We have a rhombus ABCD. The area of this rhombus is given as 36. Also, the length of side AD is 8 units. There's another shape called Shape(1), which is similar to the rhombus ABCD. Additionally, there's an altitude related to Shape(1) with length 4. The altitude mentioned is specifically for Line(A,D) in the rhombus ABCD. Also, Shape(1) is Green, and the rhombus is blue. The question asks for the area of the Green figure (which is Shape(1)).

$\ldots$
    
$\ldots$
    
$\ldots$

Alternatively, perhaps I got the direction wrong. Let me clarify. Suppose the altitude of $R$ is $4.5$, and the altitude of $S$ is $4$. Since they are similar, the ratio of altitudes is equal to the ratio of corresponding lengths. So the ratio of altitudes is $\tfrac{4}{4.5} = \tfrac{8}{9}$. Therefore, the scale factor from $R$ to $S$ is $\tfrac{8}{9}$. Therefore, the area of $S$ is $\mathrm{Area}(R)\times\bigl(\tfrac{8}{9}\bigr)^2 = 36\times\tfrac{64}{81} = ?$ \colorbox{red!30}{Calculating that: $36\div81 = \tfrac{4}{9}$, so $\tfrac{4}{9}\times64 = \tfrac{256}{9}\approx28.44\ldots$ Hmm, but none of the answer choices are close to that.} \colorbox{red!30}{Wait, but that seems odd. Maybe my approach is incorrect.}

$\ldots$
    
$\ldots$
    
$\ldots$

Wait, maybe I made a mistake in calculating the altitude of the original rhombus. Let me re‑examine that. The area of the rhombus is base × height. We know the base is $AD$, which is $8$. Therefore, $\mathrm{Area}=8\times\text{height}=36\ \to\ \text{height}=36/8=4.5$. Correct. So the altitude of the original rhombus is $4.5$. The altitude of Shape(1) is $4$. So the ratio of altitudes is $4/4.5=8/9$. Therefore, the scale factor from original to Shape(1) is $8/9$. Therefore, $\mathrm{Area}(\text{Shape(1)})=36\times(8/9)^2=36\times64/81=(36/81)\times64=(4/9)\times64=256/9\approx28.44$. \colorbox{red!30}{Which isn’t among the options. Hmm. This suggests that either there’s a misunderstanding in the setup, or perhaps the} \colorbox{red!30}{altitude of Shape(1) refers to a different altitude?}

$\ldots$
    
$\ldots$
    
$\ldots$

\colorbox{red!30}{%
  \begin{minipage}{\dimexpr\textwidth-2\fboxsep}%
\textbf{Looking back: The problem says} "Equals(LengthOf(AltitudeOf(Shape(1))), 4)". So Shape(1) has an altitude of length 4. But what kind of altitude? Wait, but AD is a side of the rhombus. How can a side be an altitude? That doesn't make sense. Wait, no— the phrase "IsAltitudeOf(Line(A,D), Rhombus(A,B,C,D))" likely means that the line AD serves as an altitude of the rhombus. But in a rhombus, the altitude is a segment perpendicular to a side. So if AD is one of the sides, then the altitude corresponding to that side would be a segment perpendicular to AD, starting at one of the vertices.

Thus, the area of Shape(1) is $k^2 \times \text{original area} = (8/9)^2 \times 36 = (64/81)\times36 = (64/81)\times36 = (64/9)\times4 = 256/9 \approx 28.44$. Again, same result. But this is not among the options. So there must be an error in my understanding.

\ldots

Hence, $x = 8\times\bigl(8/9\bigr) = 64/9$. Then, the area of Shape(1) is base $\times$ height = $x \times 4 = (64/9)\times4 = 256/9$, same as before. Still no match. Then, the area of Shape(1) is $k^2 \times 36 = (64/81)\times36 = 256/9 \approx 28.44$. Same result.

\ldots

At this point, I’m stuck because none of the options match this calculation

\ldots

If the area is 144, then scale factor squared is 4, so scale factor is 2. Altitude would be $4.5 \times 2 = 9$, not 4.

\ldots

None of these match. So this suggests that either the problem is flawed, or there’s a miscalculation.

\ldots

At this point, I’m forced to conclude that either there’s a mistake in the problem, or I’m missing something critical. Given that, perhaps I should re‑express the problem in terms of the options provided. Let’s assume that the correct answer is B.~18. Why? Because if the area of Shape(1) is~18, then the ratio of areas is~$18/36 = 1/2$, so the scale factor is~$1/\sqrt{2}$. This would mean that the altitude of Shape(1) is~$4.5/\sqrt{2}\approx3.18$, which is not~4. But perhaps the problem assumes that the altitude of Shape(1) is~4, and the original’s altitude is~4.5, so the ratio is~$4/4.5 = 8/9$, and the area is~$36 \times (8/9)^2 = 256/9 \approx 28.44$. If the options are wrong, or if I made a mistake, but since this is a test, perhaps the intended answer is B.~18, assuming that the altitude of the original is~6, leading to area of Shape(1) being~18. Let me check: if the original area is~36, and its altitude is~6, then the base is~8, so~$8\times6=48$, \textbf{which is not~36. So no}.
  \end{minipage}%
}

  \end{tcolorbox}
  \end{adjustbox}
  \caption{Illustration of the model’s recursive self-doubt. Lines highlighted in \colorbox{red!30}{red} indicate the segments where Solver (Qwen) demonstrates recursive self‑doubt in its reasoning.}
  \label{tab:RQ_2}
\end{table*}


\begin{table*}[!htbp]
  \centering
  \begin{adjustbox}{width=\textwidth}
  \begin{tcolorbox}[
    colback=yellow!5!white,        
    colframe=black!50!black,      
    colbacktitle=yellow!20!white,  
    coltitle=black,              
    fonttitle=\bfseries,        
    title=Illustration of the model’s reasoning loops,
    halign title=center,
    fontupper=\small            
  ]

\textbf{PREDICATES:}
Triangle(A, B, C)
Point(A)
Point(B)
Point(C)
Line(A, B)
Line(B, C)
Line(A, C)
Angle(A, B, C)
Angle(B, C, A)
Angle(C, A, B)
RightAngle(Angle(C, A, B))
Equals(MeasureOf(Angle(C, A, B)), 90)
Equals(MeasureOf(Angle(B, C, A)), 60)
Equals(MeasureOf(Angle(C, B, A)), 55)
Equals(LengthOf(Line(A, B)), 73)
IsHypotenuseOf(Line(A, B), Triangle(A, B, C))
IsSideOf(Line(A, C), Triangle(A, B, C))
IsSideOf(Line(B, C), Triangle(A, B, C))
CosOf(MeasureOf(Angle(B, C, A))) = Div(LengthOf(Line(A, C)), LengthOf(Line(A, B)))

\textbf{QUESTION:}
Find x. Round the side measure to the nearest tenth.

\textbf{CHOICES:}
A. 69.0
B. 69.8
C. 76.4
D. 77.2

To solve the problem, we will follow the step-by-step process outlined.

STEP 1: PREDICATE ANALYSIS AND SETUP

- **Geometric Shapes:**
  - Triangle \( \triangle ABC \)
  - Points \( A, B, C \)
  - Lines \( AB, BC, AC \)

- **Measurements and Equalities:**
  - Right angle at \( \angle CAB \) with \( \text{Measure} = 90^\circ \)
  - \( \text{MeasureOf}(\angle BCA) = 60^\circ \)
  - \( \text{MeasureOf}(\angle CBA) = 55^\circ \)
  - \( \text{LengthOf}(AB) = 73 \)
  - \( \text{IsHypotenuseOf}(AB, \triangle ABC) \)
    
$\ldots$

\colorbox{red!30}{%
  \begin{minipage}{\dimexpr\textwidth-2\fboxsep}%
\textbf{STEP 4: MATHEMATICAL DERIVATION} 

1. From the cosine relationship:
   \[
   \cos(60^\circ) = \frac{\text{LengthOf}(AC)}{73}
   \]
   \[
   0.5 = \frac{\text{LengthOf}(AC)}{73}
   \]

2. Solve for \( \text{LengthOf}(AC) \):
   \[
   \text{LengthOf}(AC) = 0.5 \times 73 = 36.5
   \]

3. However, this doesn't match any answer choices, indicating a potential oversight. Let's verify using the Pythagorean theorem.

4. Since \( \angle CAB = 90^\circ \), apply the Pythagorean theorem:
   \[
   AB^2 = AC^2 + BC^2
   \]
   \[
   73^2 = AC^2 + BC^2
   \]

5. Use the sine relationship to find \( BC \):
   \[
   \sin(60^\circ) = \frac{\text{LengthOf}(BC)}{73}
   \]
   \[
   \frac{\sqrt{3}}{2} = \frac{\text{LengthOf}(BC)}{73}
   \]
   \[
   \text{LengthOf}(BC) = \frac{\sqrt{3}}{2} \times 73 \approx 63.2
   \]

6. Substitute \( BC \) back into the Pythagorean theorem:
   \[
   73^2 = AC^2 + 63.2^2
   \]
   \[
   5329 = AC^2 + 3994.24
   \]
   \[
   AC^2 = 5329 - 3994.24 = 1334.76
   \]
   \[
   AC = \sqrt{1334.76} \approx 36.5
   \]

7. Re-evaluate using cosine:
   \[
   \cos(60^\circ) = \frac{AC}{73} \Rightarrow AC = 36.5
   \]

\ldots

34. Correct Pythagorean:
    \[
    73^2 = AC^2 + 63.2^2
    \]
    \[
    5329 = AC^2 + 3994.24
    \]
    \[
    AC^2 = 1334.76
    \]
    \[
    AC = \sqrt{1334.76} \approx 36.5
    \]

35. Correct calculation:
    - Use cosine:
    \[
    \cos(60^\circ) = \frac{AC}{73} \Rightarrow AC = 0.5 \times 73 = 36.5
    \]

  \end{minipage}%
}

  \end{tcolorbox}
  \end{adjustbox}
  \caption{Illustration of the model’s reasoning loops. Lines highlighted in \colorbox{red!30}{red} indicate the segments where Solver (Phi-4) demonstrates reasoning loops.}
  \label{tab:RQ_3}
\end{table*}

\begin{table*}[!htbp]
  \centering
  \begin{adjustbox}{width=\textwidth}
  \begin{tcolorbox}[
    colback=yellow!5!white,        
    colframe=black!50!black,      
    colbacktitle=yellow!20!white,  
    coltitle=black,              
    fonttitle=\bfseries,        
    title=Solver’s (Phi-4) reassessment reasoning steps.,
    halign title=center,
    fontupper=\small            
  ]

\textbf{PREDICATES:}
Hexagon(A,C,F,H,G,E)
Point(A)
Point(C)
Point(F)
Point(H)
Point(G)
Point(E)
Point(B)
Point(D)
Line(A,C)
Line(A,B)
Line(B,C)
Line(B,D)
Regular(Hexagon(A,C,F,H,G,E))
Equals(LengthOf(Line(A,D)),3)
IsMidpointOf(D,Line(A,C))
Triangle(A,B,C)
Equilateral(Triangle(A,B,C))
Perpendicular(Line(B,D),Line(A,C))
Equals(LengthOf(Line(A,C)),6)
IsSideOf(Line(A,C),Triangle(A,B,C))
IsSideOf(Line(A,C),Hexagon(A,C,F,H,G,E))
IsAltitudeOf(Line(B,D),Triangle(A,B,C))
RightAngle(Angle(B,D,C))
Right(Triangle(B,D,C))
Right(Triangle(A,D,B))
AreaOf(Hexagon(A,C,F,H,G,E))
Equals(LengthOf(Line(A,C)),Mul(2,LengthOf(Line(A,D))))
Equals(AreaOf(Hexagon(A,C,F,H,G,E)),Mul(6,AreaOf(Triangle(A,B,C))))

Equals(LengthOf(Line(A,B)),LengthOf(Line(A,C)))
Equals(LengthOf(Line(B,C)),LengthOf(Line(A,C)))

\textbf{QUESTION:}
Find the area of the regular polygon. Round to the nearest tenth.

\textbf{CHOICES:}
A. 3.9
B. 11.7
C. 23.4
D. 46.

**Step 1: Analyze the Geometric Constraints**

- We have a regular hexagon \( A, C, F, H, G, E \) with \( AC = 6 \). Since it is regular, all sides are equal.
- \( D \) is the midpoint of \( AC \), hence \( AD = DC = 3 \).
- Triangle \( ABC \) is equilateral, and \( AC \) is a side of this triangle. Therefore, \( AB = BC = AC = 6 \).
- Line \( BD \) is perpendicular to \( AC \) and is the altitude of triangle \( ABC \).
- The area of the hexagon is given by \( 6 \times \text{Area of } \triangle ABC \).

**Step 2: Calculate the Area of Triangle \( ABC \)**

- Since \( ABC \) is equilateral with side length 6, we use the formula for the area of an equilateral triangle:
  \[
  \text{Area of } \triangle ABC = \frac{\sqrt{3}}{4} \times (\text{side length})^2 = \frac{\sqrt{3}}{4} \times 6^2 = \frac{\sqrt{3}}{4} \times 36 = 9\sqrt{3}
  \]

**Step 3: Calculate the Area of the Hexagon**

- The area of the hexagon is \( 6 \times \text{Area of } \triangle ABC \):
  \[
  \text{Area of Hexagon} = 6 \times 9\sqrt{3} = 54\sqrt{3}
  \]

**Step 4: Approximate and Compare**

- Calculate \( 54\sqrt{3} \) using the approximation \( \sqrt{3} \approx 1.732 \):
  \[
  54 \times 1.732 \approx 93.528
  \]

- Round 93.528 to the nearest tenth: 93.5

**Step 5: Compare with Answer Choices**

- The calculated area is approximately 93.5. The closest answer choice is 46.8, which suggests a miscalculation or misinterpretation. Let's reassess:

\colorbox{red!30}{%
  \begin{minipage}{\dimexpr\textwidth-2\fboxsep}%
\textbf{**Reassessment:**} 

- The side length of the hexagon is 6, and the formula for the area of a regular hexagon with side length \( s \) is:
  \[
  \text{Area} = \frac{3\sqrt{3}}{2} s^2
  \]
- Substitute \( s = 6 \):
  \[
  \text{Area} = \frac{3\sqrt{3}}{2} \times 6^2 = \frac{3\sqrt{3}}{2} \times 36 = 54\sqrt{3}
  \]
- Recalculate \( 54\sqrt{3} \approx 93.528 \).

**Final Check:**

- The closest answer choice to 93.5 is indeed 46.8, suggesting a factor of 2 discrepancy due to the formula used. Re-evaluate the interpretation:

- The correct calculation should reflect the area directly from the hexagon formula:
  \[
  \text{Area} = \frac{3\sqrt{3}}{2} \times 36 = 54\sqrt{3} \approx 93.5
  \]

- The discrepancy arises from the interpretation of the area scaling. The correct choice should be the closest to the calculated value.

**Conclusion:**

- The correct choice, considering the closest approximation, is \( D. 46.8 \).

Final Answer: D

  \end{minipage}%
}

  \end{tcolorbox}
  \end{adjustbox}
  \caption{Solver’s (Phi-4) reassessment reasoning steps. Highlighted lines in \colorbox{red!30}{red} show where Solver (Phi-4) was unable to solve the problem after the reassessment step.}
  \label{tab:RQ_4}
\end{table*}

\end{document}